\RequirePackage{fix-cm}
\documentclass[smallextended]{svjour3}       % onecolumn (second format)
\smartqed  % flush right qed marks, e.g. at end of proof
\usepackage{graphicx}
\usepackage{cite}
\usepackage{amsmath,amssymb,amsfonts}
\usepackage{algorithmic}

\usepackage{float}
\usepackage{graphicx}
\usepackage{textcomp}
\usepackage{hyperref}
\usepackage{xcolor}
\def\BibTeX{{\rm B\kern-.05em{\sc i\kern-.025em b}\kern-.08em
    T\kern-.1667em\lower.7ex\hbox{E}\kern-.125emX}}
    
\usepackage{listings}
\usepackage{xcolor}
\usepackage{siunitx}
\usepackage{wrapfig}

\definecolor{codegreen}{rgb}{0,0.6,0}
\definecolor{codegray}{rgb}{0.5,0.5,0.5}
\definecolor{codepurple}{rgb}{0.58,0,0.82}
\definecolor{backcolour}{rgb}{0.95,0.95,0.92}

\usepackage{hyperref}
\usepackage{color}
\definecolor{red}{rgb}{1.00,0.00,0.00}
\definecolor{blue}{rgb}{0.00,0.00,1.00}
\newcommand{\cred}[1] {\textcolor{red}{#1}}
\newcommand{\cblue}[1] {\textcolor{blue}{#1}}

\begin{document}

\title{Investigating the Importance of Shape Features, Color Constancy, Color Spaces and Similarity Measures in Open-Ended 3D Object Recognition}

%\subtitle{Do you have a subtitle?\\ If so, write it here}

%\titlerunning{Short form of title}        % if too long for running head

%\author{First Author         \and
%        Second Author %etc.
%}

\author{S. Hamidreza Kasaei$^\dagger$$^\star$,  Maryam Ghorbani, Jits Schilperoort$^\dagger$, Wessel van der Rest$^\dagger$
\thanks{$^\dagger$ Department of Artificial Intelligence, University of Groningen, PO Box 407, 9700 AK, Groningen, The Netherlands. Corresponding author: hamidreza.kasaei@rug.nl}
}
%\authorrunning{Short form of author list} % if too long for running head

\institute{S. Hamidreza Kasaei \at
          Department of Artificial Intelligence, University of Groningen, PO Box 407, 9700 AK, Groningen, The Netherlands.
       \email{hamidreza.kasaei@rug.nl}           %  \\
       \and
        Maryam Ghorbani \at
        Department of Technology and Engineering, Shahrekord University,  Shahrekord, Iran.
       \and
       Jits Schilperoort \at
        Department of Artificial Intelligence, University of Groningen, PO Box 407, 9700 AK, Groningen, The Netherlands.
        \and
       Wessel van der Rest
       Department of Artificial Intelligence, University of Groningen, PO Box 407, 9700 AK, Groningen, The Netherlands. \\
}

\date{Received: date / Accepted: date}
% The correct dates will be entered by the editor

%
\titlerunning {Open-Ended 3D Object Recognition}
\authorrunning {Hamidreza Kasaei et al.}
\maketitle

\begin{abstract}
Despite the recent success of state-of-the-art 3D object recognition approaches, service robots are frequently failed to recognize many objects in real human-centric environments. For these robots, object recognition is a challenging task due to the high demand for accurate and real-time response under changing and unpredictable environmental conditions. Most of the recent approaches use either the shape information only and ignore the role of color information or vice versa. Furthermore, they mainly utilize the $L_n$ Minkowski family functions to measure the similarity of two object views, while there are various distance measures that are applicable to compare two object views. In this paper, we explore the importance of shape information, color constancy, color spaces, and various similarity measures in open-ended 3D object recognition. Towards this goal, we extensively evaluate the performance of object recognition approaches in three different configurations, including \textit{color-only}, \textit{shape-only}, and \textit{ combinations of color and shape}, in both offline and online settings. Experimental results concerning scalability, memory usage, and object recognition performance show that all of the \textit{combinations of color and shape} yields significant improvements over the \textit{shape-only} and \textit{color-only} approaches. The underlying reason is that color information is an important feature to distinguish objects that have very similar geometric properties with different colors and vice versa. Moreover, by combining color and shape information, we demonstrate that the robot can learn new object categories from very few training examples in a real-world setting. 
% \PACS{PACS code1 \and PACS code2 \and more}
% \subclass{MSC code1 \and MSC code2 \and more}
\end{abstract}

\section{Introduction}
One of the primary goals in service robotics is to develop perception capabilities that will allow robots to interact with the environment robustly. Towards this goal, a robot must be able to recognize a large set of object categories accurately. Furthermore, in order to interact with human users, this process of object recognition cannot take more than a fraction of a second. In human-centric environments, the robot may frequently face a new object that visually can be either very similar or not similar to other categories. For example, consider apples and oranges categories: \textit{what is the difference between apples and oranges?} They both fall within the class of fruits, both are edible, have a similar spherical shape and grow on trees. Although object recognition is a typical task that is performed intuitively by human cognition, it can be quite complex when a robot has to do it. 

\begin{figure}[!t]
	\includegraphics[width=1.02\linewidth]{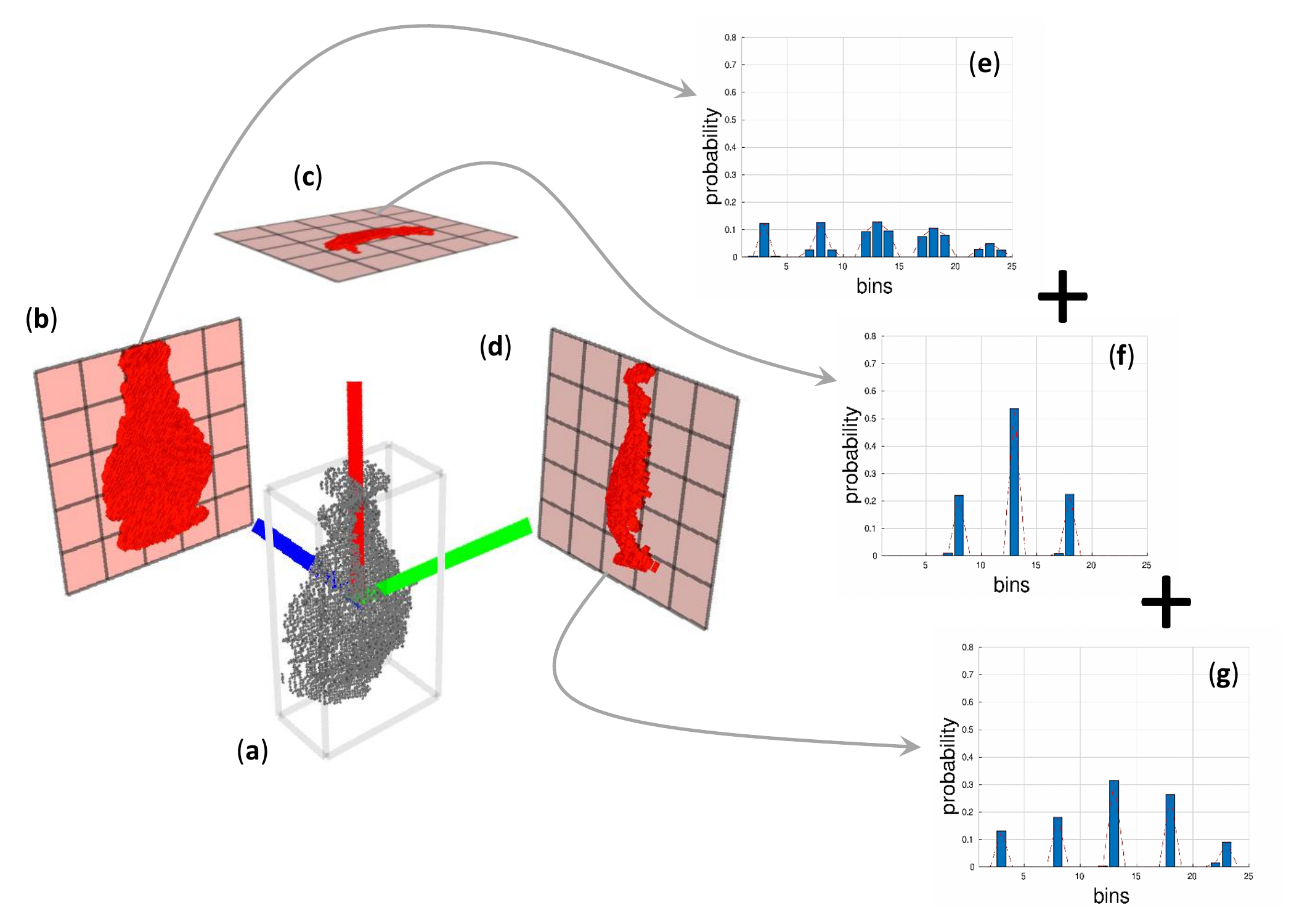}\\
	\includegraphics[width=1.1\linewidth, trim={7cm, 0cm, 0cm, 0cm, 60mm}, clip=true]{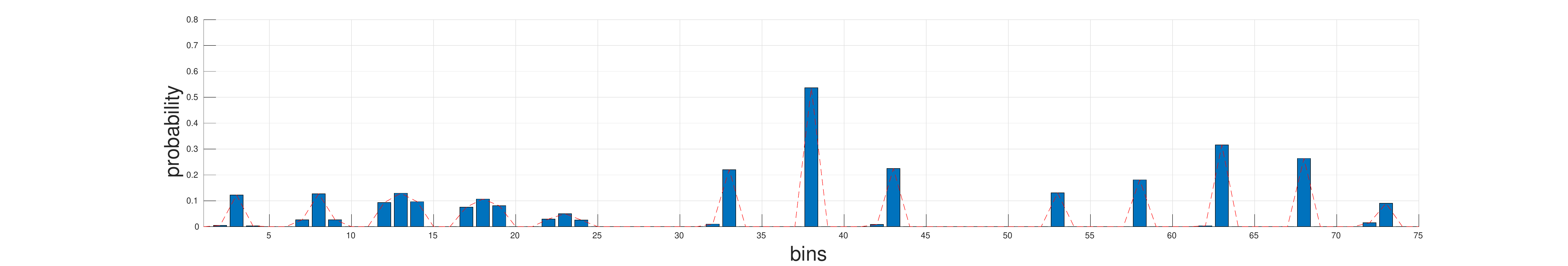}
	\caption{Step by step visualization of the process in the GOOD object descriptor, creating a representation of a `\textit{Vase}' object. (\textit{a}) shows the 3D point cloud of the object, its bounding box, reference frame, and three projected views; (\textit{b}), (\textit{c}), and (\textit{d}) show the three projection planes created from these views, i.e., the number of bins is 5. The projections are then converted into histograms by counting the number of points falling in each bin, as shown in (\textit{e}), (\textit{f}), and (\textit{g}); Finally, the GOOD object representation is created by concatenating the three histograms as visualized in (\textit{h}).}
	\label{img:good}
	\vspace{-5mm}
\end{figure}

A 3D object recognition system is composed of several software modules such as \textit{Object Detection}, \textit{Object Representation}, \textit{Object Recognition} and \textit{Perceptual Memory}. Object detection is responsible for detecting all objects in a scene. Object representation is concerned with the calculation of a set of features for the given object. The obtained representation is then sent to the object recognition module. The target object is finally recognized by comparing its representation against all the descriptions of known objects (stored in the perceptual memory). As you can see, object representation plays a prominent role because the output of this module is used for learning as well as recognition. Moreover, the representation of an object should contain enough information enabling to recognize the same or similar objects seen from different perspectives. Therefore, several important questions should be taken into account when representing an object: \textit{which perceptual data should be used? How to represent it to the robot? Which senses would a person use to classify highly similar objects?} Arguably, we can confidently state that vision would be the most important sense, while other senses such as tactile could be used for this task.

Going from this, we still do not have a definite answer on what the difference is between apples and oranges. An apple can be colored orange, while a green-colored orange could also be considered an orange. The same mutual relation goes for their shape. Taking this in mind, describing objects only by either shape or color will likely lead to confusion eventually.
In this work, we assume that an object has already been segmented from a scene. The extracted point cloud of the object, containing RGB and depth data, is used to describe the shape and color of the object for distinguishing objects that have a very similar shape with a different color or vice versa. Towards this goal, we extend the Global Orthographic Object Descriptor (GOOD) \cite{kasaei_good:_2016} by adding color constancy information as an aid to improve object recognition performance. GOOD is a light-weight object descriptor that creates a convenient object representation directly from a 3D point cloud. As 3D data contains more structural information about objects, it is more robust than RGB data to the effects of illumination and shadows \cite{regazzoni_rgb_2014}. The required steps leading to the eventual GOOD object representation for a vase object are shown in Fig.~\ref{img:good}. In summary, this paper contains the following main contributions:

\begin{itemize}
	\item Develop a 3D object descriptor that represents both shape and color constancy information for a given object.
	\item Extensively evaluate the role of shape features, color constancy, color spaces, and similarity measures in open-ended 3D object recognition.
\end{itemize}

The remainder of this paper is organized as follows. In Section~\ref{related_work}, we briefly discuss related works. The methodology for computing the object descriptor is presented in Section~\ref{approach}. Evaluation of the proposed descriptor is presented in Section~\ref{results}. Finally, in Section~\ref{conclusion}, conclusions are presented, and future research is discussed.

%%%%%%%%%%%%%%%%%%%%%%%%%%%%%%%%%%%%%%%%%%%%%%%%%%%%%%%%%%%%%%%%%%%%%%%%%%%%
%%%%%%%%%%%%%%%%%%%%%%%%%%%%%%%%%%%%%%%%%%%%%%%%%%%%%%%%%%%%%%%%%%%%%%%%%%%%
%%%%%%%%%%%%%%%%%%%%%%%%%%%%%%%%%%%%%%%%%%%%%%%%%%%%%%%%%%%%%%%%%%%%%%%%%%%%

\section{Related work}
\label{related_work}
Three-dimensional object recognition has been under investigation for a long time in various research fields, such as pattern recognition, computer graphics, and robotics\cite{martinez2017object}\cite{girshick2014rich}\cite{kim2019diversify}\cite{ullrich2017selecting}. Although an exhaustive survey of 3D object descriptors is beyond the scope of this paper~\cite{3Dreview}\cite{hana2018comprehensive}\cite{rusu20113d}, we will review the main efforts. 

Object representations based on just RGB data are sensitive to illuminations and shadows. Moreover, they cannot provide accurate representation of objects' shape. To cope with aforementioned limitations, 3D data can be used to facilitate the representation of objects. Existing 3D object representation approaches are based on either global or local descriptors. As the name suggests, global descriptors represent the complete object. In contrast, local descriptors encode an object in a piece-wise manner, representing small patches of the object around specific key points, e.g.~\cite{logoglu2016cospair}. Generally, global descriptors are increasingly used in the context of 3D object recognition, object manipulation, as well as geometric categorization. These must be efficient in terms of computation time as well as the memory, to facilitate real-time performance. Some descriptors use a Reference Frame (RF) to compute a pose invariant description. Therefore, this property can be used to categorize 3D shape descriptors into three categories, including ($i$) shape descriptors without a common reference; ($ii$) shape descriptors computed relative to a reference axis; ($iii$) shape descriptors computed relative to an RF.

Most of the shape descriptors of the first category use certain statistic features or geometric properties of the points on the surface like depth value, curvature, and surface normal to generate a description. For instance, W. Wohlkinger and M. Vincze \cite{ESF} introduced a global shape descriptor called Ensemble of Shape Functions (ESF) that does not require the use of normals to describe the object. The characteristic properties of an object are represented using an ensemble of ten 64-bin histograms of angle, point distance, and area shape functions. ESF completely ignores the potential role of color information. 

In contrast, the descriptors in the second and third category encode the spatial information of the objects’ points using a Reference Frame (RF).  In the second category, Viewpoint Feature Histogram (VFH) \cite{VFH} is a well-known descriptor. It is based on another set of descriptors, the point feature histogram (PFH)~\cite{PFH}, more specifically the fast point feature histogram (FPFH)~\cite{FPFH}. The histogram of a PFH results from considering several angular features between the normals of pairs on the point cloud. What VFH adds to FPFH is the consideration of a viewpoint component. The direction from the viewpoint to the centroid of the object is translated to all points. The angle between this and the normal of the points constitutes the first component of the histogram. The other components of the histograms are similar to FPFH, but the pan, tilt, and yaw angles are now computed between the normals of the points and the viewpoint direction of the centroid. In the third category, We have the Global Orthographic Object Descriptor (GOOD)~\cite{kasaei_good:_2016}, which performs a principal component analysis on the point cloud of an object to make an unambiguous reference frame for the object. The resulting RF is then used to create three orthogonal projection of the object with respect to the X,Y, and Z axes. Each of these projections is then converted into a histogram and then combined using two statistical features, i.e., entropy and variance, to provide the final descriptor of the object. {The Globally Aligned Spatial Distribution (GASD)~\cite{lima_efficient_2016} and VFH-Color~\cite{zrira2017vfh} are also fallen into the third category. GASD explores the idea of forming an object descriptor containing both color and shape information. GASD represents the shape information, almost similar to the GOOD descriptor. VFH-Color combines the original VFH descriptor with the color quantization histogram. Both GASD and VFH-Color descriptors incorporate color information into the descriptor to increase their discriminative power.} We refer the reader to two comprehensive surveys on local feature descriptors \cite{guo20143d,leng2018local}. In this paper, we select one descriptor from each category, including ESF, VFH, and GOOD.

{In recent studies on object recognition and grasping, much attention has been given to deep Convolutional Neural Networks (CNNs). It should be noted that there are several differences between the proposed approaches and CNN based approaches. Deep learning approaches works well if {\textit{we have a fixed set of object categories and a massive number of examples per category that are sufficiently similar to the test images}}. In real-world scenarios, these assumptions are not satisfied, and the robot needs to learn new concepts using very few training examples on-site. While deep learning is a very powerful tool, there are several limitations to use such approaches in open-ended domains. CNNs are incremental by nature but not open-ended since the inclusion of new categories enforces a restructuring in the topology of the network. In other words, the target set of classes is predefined in incremental learning, and the representation of these classes is improved over time, whereas in open-ended learning the set of classes is growing continuously. Moreover, CNNs are data-hungry approaches, and training with limited data leads to poor performance. Catastrophic forgetting is another important limitation of these approaches.}

\section{Proposed Approach}
\label{approach}
\begin{wrapfigure}{r}{0.3\textwidth}
\vspace{-2mm}
  \centering
    \includegraphics[width=0.3\textwidth]{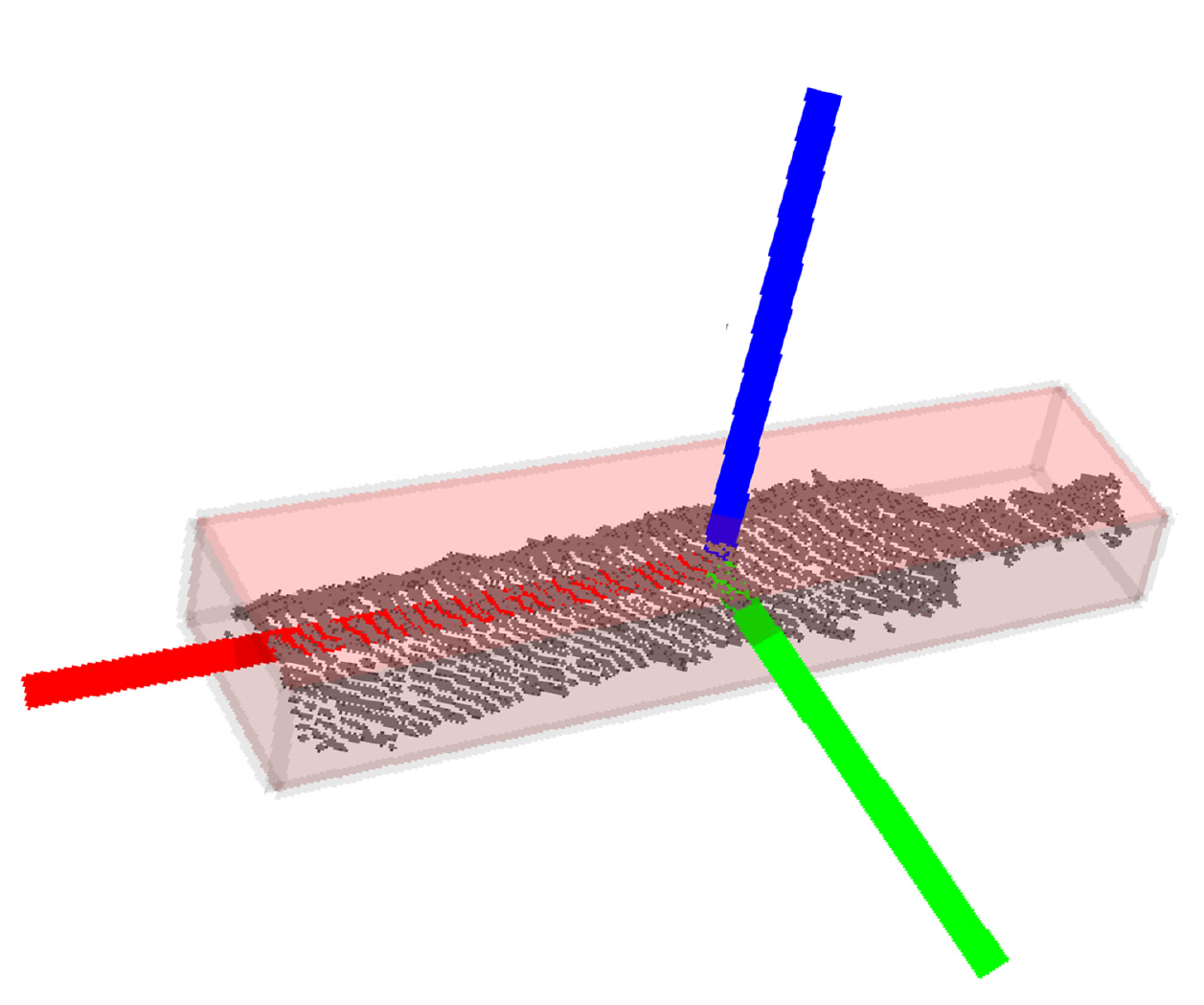}
  \caption{{Reference frame estimated for a partial view of a bottle object: the X, Y, and Z axes are represented by red, green, and blue lines. The bounding box of the object is also computed by finding minimum and maximum points in each axis.}}
  \label{reference_frame}
\end{wrapfigure}

A point cloud of an object is represented as a set of points, $\textbf{p}_i : i \in \{1,\dots,n\}$, where each point is described by their 3D coordinates $[x, y, z]$ and RGB information. In this work, we mainly use GOOD object descriptor to represent the object as a  histogram~\cite{kasaei_good:_2016}\cite{kasaei2016orthographic}. The reason why we use GOOD rather than other 3D object descriptors is that the GOOD is a pose- and scale-invariant descriptor, and therefore suitable for 3D perception in autonomous robots. As shown in Fig.\ref{img:good}, this method performs a principal component analysis on the point cloud of an object to find the eigenvectors of the object. {In particular, given a point cloud of an object that contains $n$ points, the center of gravity of the object is first calculated as $\textbf{c} = \frac{1}{n}\sum_{i=1}^{n} \textbf{p}_i$. The normalized covariance matrix, $\Sigma$, of the object is constructed:
\begin{equation}
	\Sigma = \frac{1}{n}\sum_{i=1}^{n} (\textbf{p}_i-\textbf{c})(\textbf{p}_i-\textbf{c})^T,
	\label {covariance_matrix}
\end{equation}
Then, eigenvalue decomposition is performed on the $\Sigma$, therefore, we have: $ 	\Sigma\textbf{V} = \textbf{E}\textbf{V}$
where $\textbf{V} = [\textbf{v}_1, \textbf{v}_2, \textbf{v}_3]$ contains the three eigenvectors,
$\textbf{E} = \operatorname{diag}(\lambda_1, \lambda_2, \lambda_3)$ is a diagonal matrix of the corresponding eigenvalues and 
$\lambda_1 \ge \lambda_2 \ge \lambda_3$.
Since the covariance matrix is symmetric positive, its eigenvalues are positive and the eigenvectors are orthogonal. In this work, the first two axes of object's reference frame, $X$ and $Y$, are defined by the eigenvectors $v_1$ and $v_2$, respectively. We define the $Z$ axis as the cross-product of $v_1 \times v_2$. The reference frame estimated for an example partial point cloud is shown in Fig.~\ref{reference_frame}.}
It is worth mentioning that over different trials, the direction of eigenvectors is not unique and has $\ang{180}$ ambiguity.  A sign disambiguation method is used to avoid this problem. The resulting unambiguous local reference frame, centered on the object, is then used to create three orthogonal projection planes. The projections are divided up in a grid of $n \times n$ bins, which are used to compute a normalized distribution matrix by counting how many points fall within each bin. The histogram of the plane is created by stringing the rows of the matrix together. The obtained histograms corresponding to the three projections are then combined to form a single representation for a given object. The histogram appearing first in the combined histogram is the one with the highest entropy. The second one is the one with the lowest variance of the remaining two, automatically placing the remaining one in the last position.

The GOOD object descriptor does not contain color information. Therefore, we have decided to append color constancy information to the GOOD object descriptor by taking an average color of all points of the object. The idea of considering color constancy information is inspired by the work of Bramão et al. \cite{bramao_contribution_nodate}, which showed the importance of color constancy in object recognition tasks. Therefore, the integration of color constancy information of an object seems to be sufficient to improve the performance of object recognition. Color diagnostic objects will have a single dominant color that is typical for this object and could be used for the recognition of this object. Non-color diagnostic objects will not have a dominant color value and thus can't really be used to recognize an object. Human perception and recognition, of course, do not just use color constancy information to recognize objects. However, the research by Bramão et al. \cite{bramao_contribution_nodate} showed that the color diagnosticity of an object significantly influences the performance of object recognition. In most of cases, in addition to the shape properties, it is sufficient to only look at the color constancy information. Moreover, the cost of the implementation is less than using an independent texture descriptor (e.g., ORB~\cite{rublee2011orb}), and it is not substantially altering the shape descriptor. Given this point that only $m$ bins are appended to the final object description for the color constancy information, it would not really affect the GOOD descriptor. It is worth to mention that the size of $m$ depends on the color space. In most cases, $m$ is set to three, which is much smaller than the size of the shape descriptor ($3 \times n^2$). Therefore, to avoid the dominance of the shape information, we add the parameter color weight, $w$.  {In all experiments, the obtained object representations are paired with an instance-based learning (IBL) approach (see e.g., \cite{oliveira20163d}). Therefore,  a category is described by a set of known instances. An advantage of the IBL approaches is that they can recognize objects using a minimal number of experiments, and the training phase is very fast. IBL is a baseline approach to evaluate object representations. However, more advanced approaches like SVM and Bayesian~\cite{kasaei2018perceiving}\cite{kasaei2018towards} approaches can be easily adapted. Similarly, a simple baseline recognition mechanism in the form of the nearest neighbor classifier is used. In particular, IBL approaches can be seen as a combination of particular object representation, similarity measure, and classification rule. Since representing color constancy in different color spaces may lead to different results, we convert the object's color constancy into three different color spaces, including RGB, YUV, and HSV, and further discuss the procedure of combining color constancy and shape information to form a descriptor for a given object and color spaces. In a real-world scenario, the different color spaces can lead to different results when the lighting conditions in the scene change. Therefore, additional real-world experiments are performed in which an extra light source is placed underneath the camera. The results of these experiments could show a relation between the applied color space and its robustness with regard to environmental lighting conditions.}

\textbf{ In RGB space}, often the most popular color space, colors are made up of red, blue, and green channels, having a range of values $[0, 255]$. We get the RGB values for all points of the object and calculate the summation of each channel values separately. We then get the average colors of the object by dividing the obtained red, green, blue values by the number of points of the object. Finally, since the shape information is normalized, i.e., having a range from $0$ to $1$, we also normalize the obtained color values to be in the range of $[0, 1]$, by diving each color to $255$. The obtained values are then appended to the shape description of the object. 

\textbf{YUV space} is mainly used for television transmission and represents a color by three components, one channel for luminance and two channels for chrominance. The Y component determines the brightness of color, which is referred to as luminance. The U and V component determines the color itself, also called chroma. The value of Y ranges from $0$ to $255$, while the value of U and V ranges $-128$ to $+127$. The YUV values can be derived from the RGB values using the following linear transformation:

\begin{equation}
\small
\begin{bmatrix}
Y \\
U \\
V \\
\end{bmatrix} 
=
\begin{bmatrix}
0.299 &  0.587 &  0.114  \\
-0.168 & -0.331 &  0.500 \\
0.500  & -0.418 & -0.0813 
\end{bmatrix}
\begin{bmatrix}
R \\
G \\
B \\
\end{bmatrix} 
+
\begin{bmatrix}
0 \\
128 \\
128 \\
\end{bmatrix} 
\end{equation}

\noindent where $128$ is added to $U$ and $V$ component so that each of the YUV components ranges in $[0-255]$. Afterward, the obtained colors are normalized and appended to the histogram of the object as done for the RGB color space.

\textbf{HSV color space} was developed to take into consideration how humans view color, where $H$ stands for hue, $S$ stands for saturation, and $V$ stands for value. In particular, it describes a color (hue) in terms of the saturation (shade) and value (brightness). The hue components represent the angle, and its value ranges $[0, 360]$ degree. The saturation component describes the percentage of gray in a particular color and value works in conjunction with saturation and describes the brightness or intensity of the color, range from $[0, 100]$ percent. The RGB value of every detected point is converted to HSV; this is done using the minimum and maximum value of the normalized RGB value of the point in the point cloud. It is worth to mention, the final object descriptor is formed the same way as the other two color spaces. The normalized HSV color dissimilarity, $d$, of two object views, $p$ and $q$, can be computed using the following equations:

\begin{equation}
h = \frac{min(abs(h_q - h_p),~360-abs(h_q-h_p)}{180.0}\\
\nonumber
\end{equation}
\begin{equation}
s = abs(s_q-s_p)
\nonumber
\end{equation}
\begin{equation}
v = \frac{abs(v_q-v_p)}{255.0}
\end{equation}
\begin{equation}
d = h + s + v
\nonumber
\end{equation}

{As stated earlier, after forming the object descriptor containing both color and shape information, we use the \textit{color weights} parameter to set how important the difference in color is of the two compared object representations. We are doing this because in the representation of the object, the number of bins representing the shape of the object is much more than the number of bins representing the color information (675 bins vs. 3 bins), and hence the shape information will largely dominate the decision. We, therefore, calculate the difference by using a weighted distance function, as shown below: 
\begin{equation} \label{eq:color_weight}
D (p, q) = (1-w) \times d_{s}(p, q) + w \times d_{c}(p, q)
\end{equation}
\noindent where $d_{s}$ is the difference in the shape space, $d_{c}$ is the difference in the color space, and $w$ is the color weight, which is a value between $0.0$ and $1.0$.}

\section{Result and Discussion}
\label{results}
Three types of experiments were carried out to evaluate the proposed approach.  It should be noted that in addition to GOOD descriptor~\cite{kasaei_good:_2016}, two popular state-of-the-art 3D object descriptors including, VFH~\cite{VFH} and ESF~\cite{ESF} were evaluated, which are available in the Point-Cloud Library\footnote{\href{http://pointclouds.org/}{http://pointclouds.org/}}. We compare the obtained results and use the best configuration as default system's configuration in the second round of experiments (open-ended evaluation). In the following subsections, we have investigated the importance of shape information and similarity measures using an extensive set of offline evaluations and considered the importance of color constancy and color spaces in a broad set of open-ended assessments.

\subsection{Classical offline evaluation using restaurant object dataset}

\begin{table}[!b]
	\begin{center}
		\caption {\small System configurations that obtained \textit{best} object recognition accuracy}
		\vspace{-3mm}
		\begin{tabular}[t]{|c|c|c|c|c|c|}
		\hline
		\textbf{No.} &	\textbf{Descriptor} & \textbf{\#bins}& \textbf{Distance Function} & \textbf{Accuracy} & \textbf{{Time(s)}}\\
			\hline
			1 & GOOD	&   15	& Manhattan	& 0.97 & 3.351\\\hline
			2 & GOOD	&	15	&	Gower	&	0.97  & 3.450 \\\hline
			3 & GOOD	&	15	&	Sorensen&	0.97  & 3.512\\\hline
			4 & GOOD	&	15	&	Motyka		& 0.97 & 3.666 \\\hline
			5 & GOOD	&	15	&	Euclidean	 &	0.97  & 4.078 \\\hline
			6 & GOOD	&	15	&	Cosine	&	0.97  & 4.297 \\\hline
			7 & GOOD	&	15	&	Dice	&	0.97  & 4.380 \\\hline
			8 & GOOD	&	25	&	$\chi^2$	&0.97  & 4.219\\\hline
			9 & GOOD	&	30	&	Bhattacharyya&	0.97  & 5.771\\\hline
			10 & GOOD	&	35	&	Bhattacharyya&	0.97  & 6.116\\\hline
			11 & GOOD	&	35	& 	$\chi^2$	&	0.97  & 5.348\\\hline
			12 & ESF	&	---	&	Manhattan	&	0.97  & 8.055\\\hline
			13 & ESF	&	---	&	Sorensen	&	0.97  & 7.983\\\hline
			14 & ESF	&	---	&	Neyman	&	0.97  & 7.983 \\\hline
            15 & ESF	&	---	&	Bhattacharyya&	0.97  & 8.055\\\hline
			16 & ESF	&	---	&	Euclidean	&	0.97  & 8.174\\\hline
			
		\end{tabular}
		%\vspace{-5mm}
		\label{table1}
	\end{center}
\vspace{-4mm}
\end{table}

For this round of experiments, we have used the restaurant object dataset since it has a small number of classes (10 categories) with a significant intra-class variation that is suitable for performing extensive sets of experiments. The parameter of the selected object descriptors must be tuned to provide a good balance between recognition performance, memory usage, and processing speed. The descriptiveness of the GOOD descriptor was evaluated with varying \textit{number of bins}, $n$, ranging from $5$ to $50$ with the interval of $5$. For the VFH descriptor, we performed a parameter sweep on the \textit{normal estimation radius} parameter, ranging from $2cm$ to $10cm$ with the interval of $2cm$, to find the value which resulted in the highest accuracy. The ESF object descriptor does not have any parameters to be optimized. Furthermore, the choice of the similarity
measure has an impact on the recognition performance. In the case of the similarity measure, since the selected object descriptors represent an object as a normalized histogram, the dissimilarity between two histograms can be computed by different distance functions. We refer the reader to a comprehensive survey on distance/similarity measures provided by S. Cha \cite{cha2007comprehensive}. In this work, during the selection of the distance functions, care was taken to select functions that were dissimilar from each other. This policy will increase the chance that different distance functions lead to different results. Based on these considerations, the following 14 functions have been explored: \textit{Euclidean, Manhattan, $\chi^2$, Pearson, Neyman, Canberra, KL divergence, symmetric KL divergence, Motyka, Cosine, Dice, Bhattacharyya, Gower,} and \textit{Sorensen}. We refer the reader to \cite{cha2007comprehensive} to check the mathematical equations. We therefore performed a total of $224 = (10\times14)+ (5\times14) + 14$  ten fold cross-validation experiments to obtain best configuration for each method. All combination of parameters that obtained the \textit{best accuracy} is summarized in Table \ref{table1}. By comparing all experiments, it is clear that GOOD with $15$ bins and Manhattan (city-block) distance function configuration obtained the best performance in terms of \textit{accuracy} and \textit{computation time}. The Manhattan distance function is in the $L_p$ Minkowski family and has very low computational expenses. The accuracy of the proposed system with this configuration was $0.97$. A complete experiment (including both learning and recognition phases) using this configuration took $3.351$ seconds. The following results are computed using this configuration unless otherwise noted.

 Although a large number of bins provides more details about the point distribution, it increases computation time, memory usage, and sensitivity to noise. The descriptiveness of VFH was not as good as the other descriptors. VFH with the radius parameter set to $6cm$ and \textit{Canberra} distance function resulted in the best performance with a $0.94$ accuracy followed by the same radius parameter and \textit{Motyka} function which resulted in an accuracy of $0.93$. One crucial observation is that for {VFH}, there is a significant drop in performance when the normal estimation radius becomes too small or too large. It was observed that ESF performed well on all distance functions, always having a precision greater than $0.95$.

{Several factors influence the computation time of an instance-based object recognition approach. They are including (\textit{i}) number of training instances, (\textit{ii}) distance function, and (\textit{iii}) object descriptor. In this round of experiments, the number of instances was the same for all the approaches. In the case of distance function, it is worth mentioning that a distance function requires the same amount of time regardless of the input objects, i.e., constant time O(1). Therefore, distance functions should be compared in terms of mathematical complexity and not time complexity. By comparing the mathematical equation of the selected distance functions, it is clear that the Manhattan function is the optimum one. To evaluate an object descriptor's computation time, we randomly select 20 objects and measure the average computation time required to generate representations for the given objects. We observed that GOOD achieved the best performance, which was around $10$ and $44$ times better than ESF and VFH, respectively. The underlying reason is that GOOD works directly on 3D point clouds and requires neither triangulation of the object's points nor surface meshing \cite{kasaei2016orthographic}. Therefore, we use the GOOD descriptor as the basis of the proposed model in the remaining experiments.}

\subsection{Open-ended evaluation using RGB-D object dataset}
In this round of experiments, we explore the importance of color constancy and color spaces. To evaluate the performance of object recognition approaches in an open-ended domain, Kasaei et al. \cite{kasaei_interactive_2015} has recently adopted a teaching protocol which simulated the simultaneous nature of learning and recognition. The main idea is to emulate the interactions of a robot with the surrounding environment over long periods. The teaching protocol determines which examples are used for training the algorithm, and which are used to test the algorithm. This protocol is based on a Test-then-Train scheme, which can be followed by a human user or by a simulated user. We develop a simulated teacher to follow the protocol and autonomously interacts with the system using \textit{\textbf{teach}}, \textit{\textbf{ask}} and \textit{\textbf{correct}} actions. In this experiment, the robot initially has zero knowledge, and the training instances become gradually available according to the teaching protocol. 

The idea is that the simulated teacher introduces a category to the robot using three randomly selected object views. The robot creates a model for that category based on these instances. Afterward, the teacher picks a never-seen-before object view and tests the robot to see if it has learned the category, and learning this category does not interfere with the previously learned categories. This is done by asking the robot to recognize unseen object views of the currently known categories. When the robot makes a misclassification, the teacher will provide feedback with the correct category. This way, the robot adjusts its category model using the mistaken instance. The simulated teacher estimates the recognition accuracy of the robot using a sliding window of size $3n$ iterations, where $n$ is the number of categories that the robot has already learned. If the number of iterations it took since the last time the agent learned a new category is less than $3n$, all results are used. If the recognition performance of the agent is higher than the protocol threshold, $\tau$, the simulated teacher introduces a new category. {It is worth mentioning that the original protocol suggests to set the $\tau$ to $0.67$, meaning the object recognition accuracy is at least twice the error rate. In our experiment, we set the protocol threshold to $80\%$ since we aim to force the robot to learn and recognize object categories more precise. This way, object recognition accuracy is at least four times better than error. Furthermore, considering such a high threshold not only makes it harder to learn new object categories, but also highlights the importance of combining color and shape information in open-ended learning scenario. This relatively high protocol threshold also allows for robustness tests, as configurations that are still able to learn many categories can be considered to be more robust.} If the agent could not meet this protocol threshold after a certain number of iterations (e.g., $100$), a breakpoint is encountered. This way, the simulated teacher can state that the agent can not learn any more categories. The agent may learn all existing categories before reaching to the breaking points. In such cases, it is no longer possible to continue the protocol, and the evaluation process is halted. In the reported results, this is shown by the stopping condition, ``\textit{lack of data}''. 
%We show that color spaces can significantly affect classification accuracy in open-ended settings. 

\subsubsection{Dataset and evaluation metrics} 
In this round of experiments, we use the Washington RGB-D dataset~\cite{RGBD_dataset}. This dataset is known as one of the largest available 3D objects datasets and consists of 51 categories with 250.000 views of 300 objects. When an experiment is carried out, learning performance is evaluated using several measures \cite{oliveira20163d}\cite{lopes2007many}\cite{kasaei2018coping}, including: (\textit{i}) the number of learned categories (NLC) at the end of the experiment, an indicator of \textit{how much the system was capable of learning}; (\textit{ii}) the number of question/correction iterations (QCI) required to learn those categories and the average number of stored instances per category (AIC), indicators of \textit{time and memory resources required for learning}; (\textit{iii}) Global Classification Accuracy (GCA), computed using all predictions in a complete experiment, and the Average Protocol Accuracy (APA), i.e. average of all accuracy values successively computed to control the application of the teaching protocol. GCA and APA are indicators of \textit{how well the system learns}.

\begin{table}[!b]
	\begin{center}
		\caption {\small Summary of evaluation using shape information}
		\vspace{-2mm}
		\resizebox{\linewidth}{!}{	
			\begin{tabular}[t]{|c|c|c|c|c|c|c|c|c|}
				\hline
				No.  &  $w$  &  QCI  &  NLC  &  AIC  &  GCA  &  APA\\\hline
				1	 &	0.0  &	648.10 $\pm$ 196.76	& 18.90 $\pm$ 4.38	&11.78 $\pm$ 1.35	&0.74 & 0.84 \\\hline
		\end{tabular}}
		%\vspace{-5mm}
		\label{shape_only}
	\end{center}
	\vspace{-4mm}
\end{table}

\begin{table}[!b]
	\begin{center}
		\caption {\small Summary of evaluation in RGB color space}
		\vspace{-2mm}
		\resizebox{\linewidth}{!}{	
			\begin{tabular}[t]{|c|c|c|c|c|c|c|c|c|}
				\hline
				No.  &  $w$  &  QCI  &  NLC  &  AIC  &  GCA  &  APA\\\hline
				1  &  0.1  &  922.20 $\pm$ 459.24  &  24.10 $\pm$ 7.53  &  11.60 $\pm$ 1.84  &  0.76  &  0.84 \\\hline
				2  &  0.2  &  1217.70 $\pm$ 669.51  &  31.80 $\pm$ 11.66  &  10.84 $\pm$ 1.40  &  0.78 &  0.84 \\\hline
				3  &  0.3  &  1881.60 $\pm$ 555.00  &  44.70 $\pm$ 8.10  &  11.28 $\pm$ 1.27  &  0.80  &  0.841\\\hline
				4  &  0.4  &  1751.80 $\pm$ 477.00  &  45.70 $\pm$ 7.04  &  10.19 $\pm$ 1.19  &  0.81 &  0.85 \\\hline
				5  &  0.5  &  1656.20 $\pm$ 260.22  &  49.40 $\pm$ 4.72  &  8.92 $\pm$ 0.80  &  0.82  &  0.85\\\hline
				\cred{${\textbf{*}}$}6  &  0.6  &  1632.50 $\pm$ 153.28  & \textbf{ 51.00 $\pm$ 0.0}  &  8.30 $\pm$ 0.77  &  0.84 &  0.86 \\\hline
				\cred{${\textbf{*}}$}7  &  0.7  &  1509.50 $\pm$ 104.62  &  \textbf{ 51.00 $\pm$ 0.0}  &  7.55 $\pm$ 0.57  &  0.85  &  0.86 \\\hline
				\cred{${\textbf{*}}$}8  &  0.8  &  1452.30 $\pm$ 76.15  &  \textbf{ 51.00 $\pm$ 0.0}  &  7.07 $\pm$ 0.47  &  \textbf{0.86} &  0.87 \\\hline
				\cred{${\textbf{*}}$}\cblue{\textbf{9}}  &  \cblue{\textbf{0.9}}  &  \cblue{\textbf{1410.20 $\pm$ 43.18}}  & \cblue{\textbf{ 51.00 $\pm$ 0.0}}  &  \cblue{\textbf{6.79 $\pm$ 0.38}}  &  \cblue{\textbf{0.86}}  &  \cblue{\textbf{0.88}}\\\hline
				10  &  1.0  &  1257.10 $\pm$ 609.35  &  33.30 $\pm$ 10.84  &  10.50 $\pm$ 1.38  &  0.79  &  0.85\\\hline
		\end{tabular}}
		%\vspace{-5mm}
		\label{table2_RGB}
		\\\scriptsize{\cred{$^{\textbf{(*)}}$} Stopping condition was ``\emph{lack of data}''. Best result highlighted by blue color.}
	\end{center}
	\vspace{-4mm}
\end{table}

\subsubsection{Results}
\begin{figure}[!t]
	\begin{center}	
		\includegraphics[width=0.7\linewidth, trim={0cm, 0cm, 0cm, 0cm, 0mm}, clip=true]{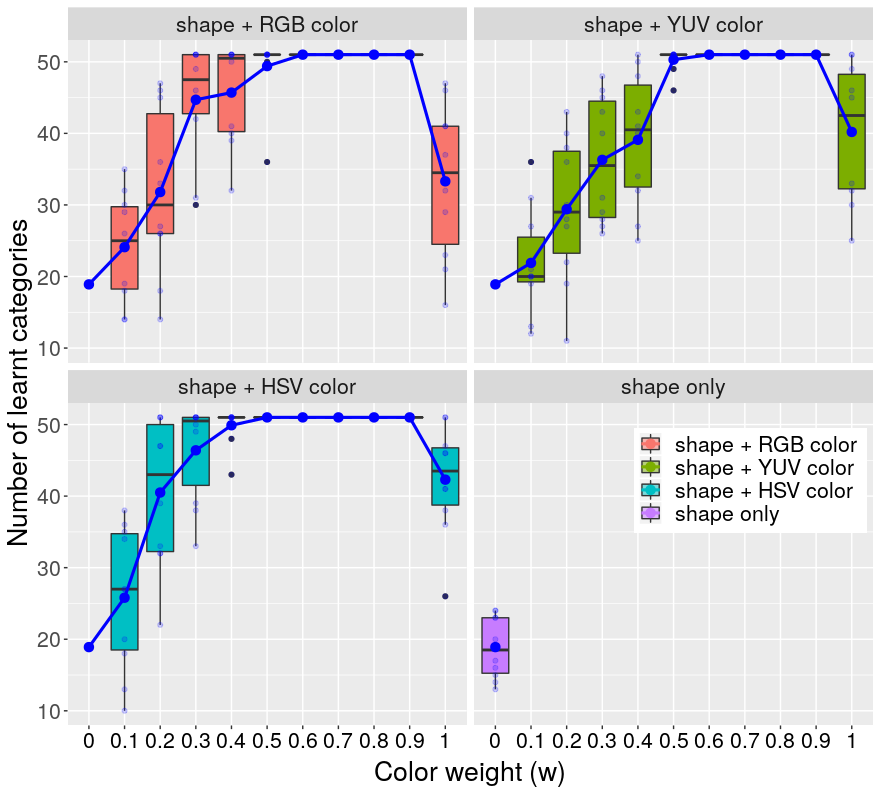}
	\end{center}
	\caption{Summary of open-ended evaluation of all approaches; These plots show the number of learned categories versus color weight for all experiments in four different space. Boxplots represent the distribution of obtained results for each configuration based on \textit{minimum}, \textit{first quartile}, \textit{median}, \textit{third quartile}, and \textit{maximum} performances. The blue lines represent the \textit{average} number of learnt categories as a function of color weight.}
	\vspace{-2mm}

	\label{NLC_W}
\end{figure}

\begin{figure}[t]
	\begin{center}	
		\includegraphics[width=0.7\linewidth, trim={0cm, 0cm, 0cm, 0cm, 0mm}, clip=true]{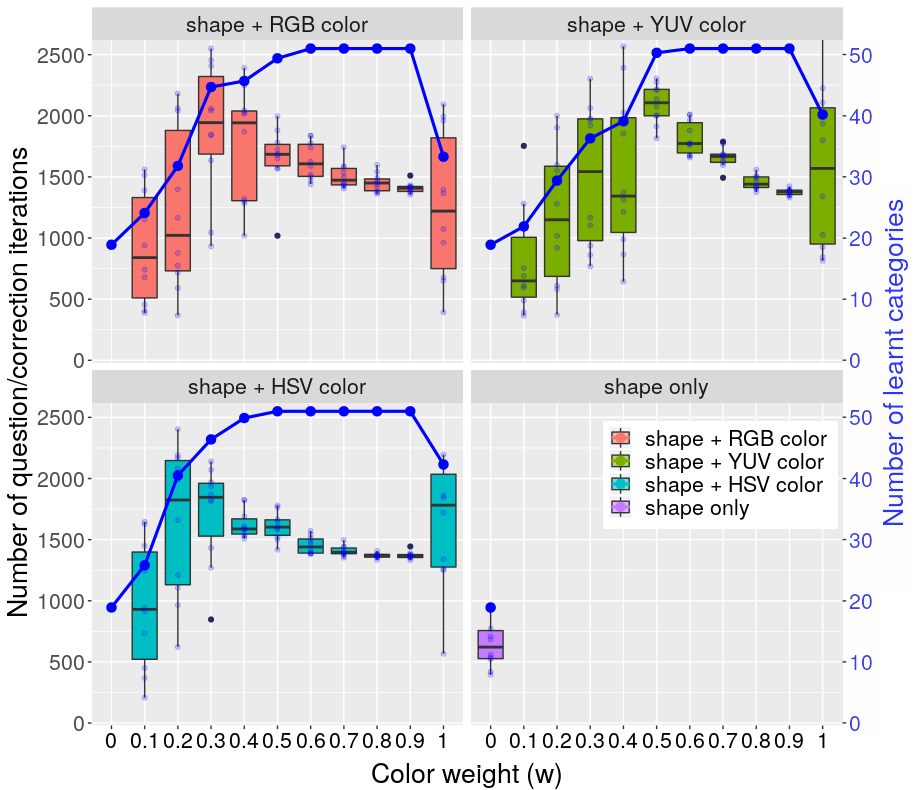}
	\end{center}
	\caption{Summary of open-ended evaluations: these graphs show the number of question/correction iterations (QCI) required to learn a certain number of categories as a function of color weight. The blue lines also represent the average number of learned categories in different combinations of color and shape.}
	\label{QCI_w}
	\vspace{-5mm}
\end{figure}

\begin{figure}[!b]
	\includegraphics[width=1\linewidth, trim={7cm, 0.75cm, 6.2cm, 0cm}, clip=true]{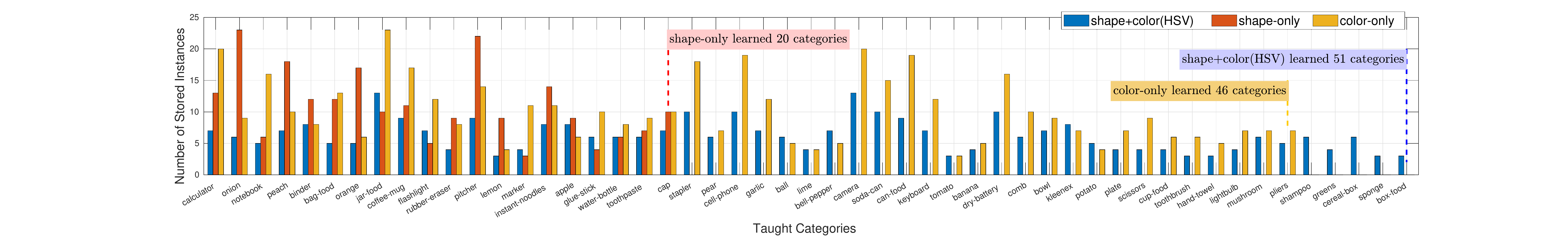}
	\caption{This graph shows the number of instances stored in the models of all of the categories in three system configurations: shape-only, color-only, and shape+HSV ($w=0.9$). Each bar represents the three instances provided at the introduction of the category, together with any instances that had to be corrected somewhere along the experiment. \textit{Onion}, \textit{jar-food}, and \textit{camera}  were the most difficult categories for \textit{shape-only}, \textit{color-only} and \textit{shape+color} configurations respectively, i.e., requiring the largest number of instances. It should be noted that categories that were introduced near the end of the experiment have been tested less, which is clearly visible in a general trend for fewer instances to be included for categories appearing later. The agent learned $20$, $46$, and $51$ categories with shape-only, color-only, and shape+color (HSV) configurations respectively. It is worth to mention that the shape+color experiment finished due to \textit{lack of data} condition, showing the potential to learn many more categories.}
	\label{NSIC}
\end{figure}

\begin{figure}[!t]
	\begin{center}	
		\includegraphics[width=0.7\linewidth, trim={0cm, 0cm, 0cm, 0cm, 0mm}, clip=true]{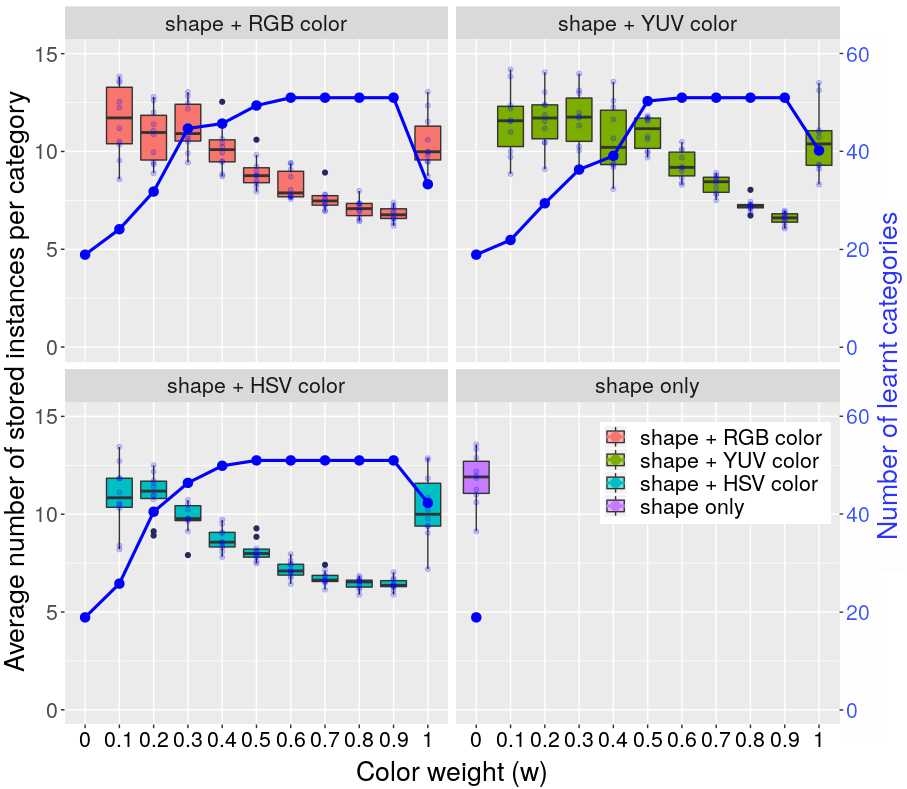}
	\end{center}
	\caption{Summary of open-ended evaluations: these graphs represent the average number of stored instances per category and the average number of learned categories at the end of experiments as an indicator of how much memory does each approach take to learn a certain number of categories. The blue lines display the average number of learned categories as a function of color weight.}
	\label{NSI_W}
	%\vspace{-2mm}
\end{figure}

\begin{figure}[!b]
	\begin{center}
		\includegraphics[width=\linewidth, trim={0cm, 0cm, 0cm, 0cm, 0mm}, clip=true]{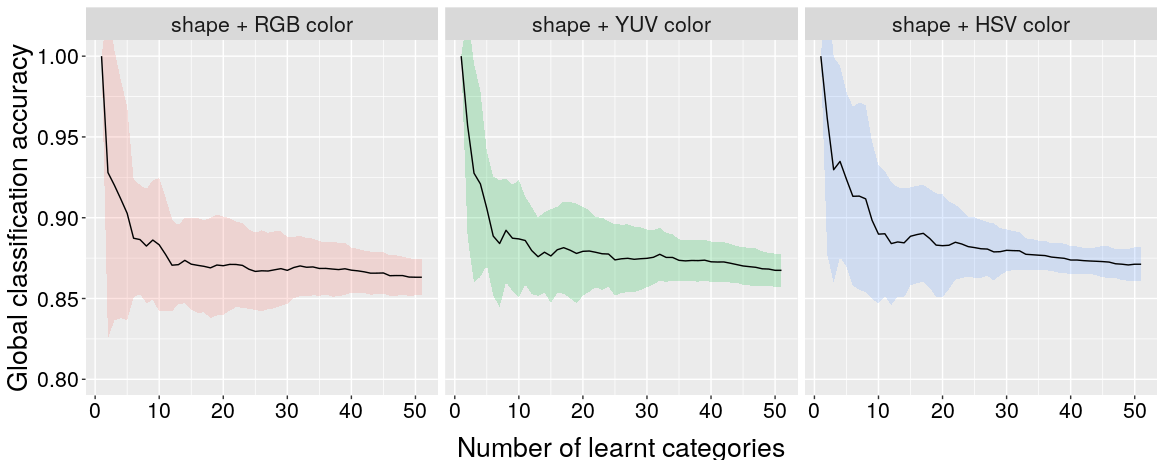}
	\end{center}
	\caption{These graphs show the global classification accuracy as a function of the number of learned categories in three different color spaces. In all these experiments, color weight was set to $0.9$.}
	\label{GCA_NLC}
	\vspace{-3mm}
\end{figure}

Since the order of introducing the categories may have an effect on the performance of the system, ten experiments were carried out for each of \textit{shape-only}, \textit{color-only} ($w=1.0$),  and \textit{nine combinations of shape and color} in three mentioned color spaces, i.e., $w \in \{0.1, 0.2, \dots, 0.9\}$, resulting $330$ experiments. This is due to the nature of IBL approaches that the recognition of new objects relies on all the previously learned objects. For example, if the teacher introduces a red apple right after a red tomato (both a red color and a similar shape), it would be harder to recognize this new object than when a banana followed the red tomato (different color and different shape) are introduced. Detailed summaries of the obtained results are reported in Tables \ref{shape_only}~--~\ref{table2_YUV}, and depicted in Figures ~\ref{NLC_W}~--~\ref{GCA_NLC}. For all results, boxplots are added to show the variation of obtained results for each configuration based on \textit{minimum}, \textit{first quartile}, \textit{median}, \textit{third quartile}, and \textit{maximum} performances. Line plots are also added to display the \textit{average} number of learned categories as a function of color weight.

\begin{table}[t]
	\begin{center}
		\caption {\small Summary of evaluation in YUV color space}
		\vspace{-2mm}
		\resizebox{\linewidth}{!}{
		\begin{tabular}[t]{|c|c|c|c|c|c|c|c|c|}
		\hline
				No.  &  $w$  &  QCI  &  NLC  &  AIC  &  GCA  &  APA\\\hline
			1  &  0.1  &  802.30 $\pm$ 443$\pm$21  &  21.90 $\pm$ 7.49  &  11.53 $\pm$ 1.72   &  0.75   &  0.84 \\\hline
			2  &  0.2  &  1183.20 $\pm$ 572.72  &  29.40 $\pm$ 10.13  &  11.64 $\pm$ 1.43  &  0.77   &  0.84 \\\hline
			3  &  0.3  &  1507.00 $\pm$ 587.95  &  36.30 $\pm$ 8.87  &  11.62 $\pm$ 1.53  &  0.79   &  0.84 \\\hline
			4  &  0.4  &  1524.10 $\pm$ 655.25  &  39.10 $\pm$ 9.31  &  10.65 $\pm$ 1.79  &  0.80   &  0.84 \\\hline
			5  &  0.5  &  2095.50 $\pm$ 161.95  &  50.30 $\pm$ 1.64  &  10.95 $\pm$ 0.85  &  0.81  &  0.84 \\\hline
			\cred{${\textbf{*}}$}6  &  0.6  &  1817.70 $\pm$ 138.34  & \textbf{ 51.00 $\pm$ 0.0}  &  9.31 $\pm$ 0.76  &  0.82   &  0.85 \\\hline
			\cred{${\textbf{*}}$}7  &  0.7  &  1659.90 $\pm$ 84.90  &  \textbf{51.00 $\pm$ 0.0}  &  8.32 $\pm$ 0.47  &  0.84 &  0.86 \\\hline
			\cred{${\textbf{*}}$}8  &  0.8  &  1455.10 $\pm$ 58.64  & \textbf{ 51.00 $\pm$ 0.0 } &  7.23 $\pm$ 0.35  &  0.85  &  0.87 \\\hline
			\cred{${\textbf{*}}$}\cblue{\textbf{9}}  &  \cblue{\textbf{0.9}}  &  \cblue{\textbf{1375.50 $\pm$ 26.95}}  & \cblue{\textbf{51.00 $\pm$ 0.0}}  &  \cblue{\textbf{6.58 $\pm$ 0.31}}  &  \cblue{\textbf{0.87 }}  & \cblue{\textbf{ 0.88 }}\\\hline
			10  &  1.0  &  1568.30 $\pm$ 664.25  &  40.20 $\pm$ 9.56  &  10.55 $\pm$ 1.71  &  0.80   &  0.84 \\\hline
		\end{tabular}}
		%\vspace{-5mm}
		\label{table2_YUV}
		\\\scriptsize{\cred{$^{\textbf{(*)}}$} Stopping condition was ``\emph{lack of data}''. Best result highlighted by blue color.}
	\end{center}
	\vspace{-4mm}
\end{table}

\begin{table}[t]
	\begin{center}
		\caption {\small Summary of evaluation in HSV color space}
		\vspace{-2mm}
		\resizebox{\linewidth}{!}{
		\begin{tabular}[t]{|c|c|c|c|c|c|c|c|c|}
			\hline
				No.  &  $w$  &  QCI  &  NLC  &  AIC  &  GCA  &  APA\\\hline
			1  &  0.1  &  958.80 $\pm$ 523.49  &  25.80 $\pm$ 10.08  &  10.89 $\pm$ 1.69  &  0.77  &  0.84 \\\hline
			2  &  0.2  &  1639.30 $\pm$ 618.87  &  40.50 $\pm$ 10.33  &  11.01 $\pm$ 1.18  &  0.80   &  0.84 \\\hline
			3  &  0.3  &  1717.40 $\pm$ 407.80  &  46.40 $\pm$ 6.92  &  9.90 $\pm$ 0.98  &  0.81   &  0.84 \\\hline
			4  &  0.4  &  1628.00 $\pm$ 114.81  &  49.90 $\pm$ 2.60  &  8.7079 $\pm$ 0.62  &  0.83   &  0.85 \\\hline
			\cred{${\textbf{*}}$}5 &  0.5  &  1608.50 $\pm$ 113.54  &  \textbf{51.00 $\pm$ 0.0} &  8.12 $\pm$ 0.56  &  0.84   &  0.86 \\\hline
			\cred{${\textbf{*}}$}6  &  0.6  &  1454.20 $\pm$ 71.34  &  \textbf{51.00 $\pm$ 0.0} &  7.17 $\pm$ 0.45  &  0.85 &  0.87 \\\hline
			\cred{${\textbf{*}}$}7  &  0.7  &  1406.20 $\pm$ 43.20  & \textbf{ 51.00 $\pm$ 0.0} &  6.73 $\pm$ 0.36  &  \textbf{0.87 }  &  0.88 \\\hline
			\cred{${\textbf{*}}$}8  &  0.8  &  \textbf{1369.60 $\pm$19.66}  &  \textbf{51.00 $\pm$ 0.0}  &  6.46 $\pm$ 0.29  &  \textbf{0.87}  &  0.88\\\hline
			\cred{${\textbf{*}}$}\cblue{\textbf{9}}  &  \cblue{\textbf{0.9}}  &  \cblue{\textbf{1371.90 $\pm$ 28.90}}  &  \cblue{\textbf{51.00 $\pm$ 0.0}}  &  \cblue{\textbf{6.42 $\pm$ 0.33}}  &  \cblue{\textbf{0.87}}  &  \cblue{\textbf{0.89 }}\\\hline
			10  &  1.0  &  1624.00 $\pm$ 513.38  &  42.30 $\pm$ 7.66  &  10.34 $\pm$ 1.78  &  0.81  &  0.85\\\hline	
		\end{tabular}}
		%\vspace{-5mm}
		\label{table2_HSV}
		\\\scriptsize{\cred{$^{\textbf{(*)}}$} Stopping condition was ``\emph{lack of data}''. Best result highlighted by blue color.}
	\end{center}
	\vspace{-4mm}
\end{table}

One important observation is that considering color constancy information significantly improved object recognition performance. It was found that the performance of the agent is improved by increasing the level of color weight in all color spaces. Notably, the agent learned all 51 categories in all color spaces when the color weight was in the range of $0.6 \le w \le 0.9$. It is worth to mention, in this range, all experiments concluded prematurely due to the ``\textit{lack of data}'', i.e., no more categories available in the dataset, indicating the potential for learning many more categories. Moreover, it was observed that the agent with neither \textit{color-only} nor \textit{shape-only} configurations could learn all categories in all of the experiments.

On closer inspection, we can see that the combination of HSV color and shape model resulted in a better performance in all levels of color combination, as clearly shown in Fig.~\ref{NLC_W}. By comparing all approaches, it is also visible that the agent learned all categories faster in HSV space than in other color spaces. It can also be concluded that shape+HSV ($w = 0.9$) obtained the best GCA and APA with stable performance. In contrast, the performance of the agent with \textit{shape-only} ($w = 0.0$) configuration was the worst among the evaluated configurations. In the case of \textit{color-only} ($w = 1.0$), the best performance was obtained in HSV color space, where the agent on average learned $42.30$ categories, and YUV and RGB spaces achieved the second and third places by learning on average $40.20$ and $33.30$ categories respectively.

Fig.~\ref{QCI_w} illustrates \textit{``how fast''} the learning occurred in each of the experiments while shedding light on the number of learned categories ({blue lines}). It shows the number of question/correction iterations (QCI) required to learn a certain number of categories. We can see that, on average, the longest experiments were observed with shape+YUV, when the $w$ parameter was set to $0.5$. The shortest ones were observed with shape+HSV with $w = 0.8$. It should be noted that the agent with shape+HSV ($w = 0.8$) configuration was able to learn all 51 categories in all experiments, while the experiments with shape+YUV ($w = 0.5$) were stopped due to reaching the \textit{break point} condition after leaning 50 categories on average (see Table \ref{table2_HSV} and \ref{table2_YUV}). In
the case of shape+RGB, the best performance of the agent was achieved when the $w$ set to $0.5$. With this {shape+RGB} configuration, the agent on average learned all categories using $1410.20\pm43.18$ question/correction iterations. It was also observed that the longest experiments were continued for $1881.60\pm555.00$ question/correction iterations with {shape+RGB} ($w=0.3$) configuration and the agent on average was able to learn
$44.70\pm8.10$ categories (see Table \ref{table2_RGB}).

Fig.~\ref{NSIC} represents the exact number of stored instances per category for shape-only, color-only (HSV), and shape+HSV ($w=0.8$). By comparing the obtained results, it can be concluded that the agent with shape+HSV configuration not only stored much fewer instances per category but also it could learn more categories as well. 
Fig.~\ref{NSI_W} provides a detailed summary of the obtained results concerning the average number of stored instances per category (AIC) as a function of color weight. By comparing all approaches, it is clear that {shape+HSV}, {shape+YUV}, and {shape+RGB} on average stored less than seven instances per category to learn all categories, while \textit{shape-only} and \textit{color-only} required more than 10 instances per categories to learn $18.90$ and $33.30$ categories respectively. The {shape+HSV} ($w = 0.9$) configuration on average stored smallest number of instances per category (see Table~\ref{table2_HSV}).

\begin{figure}[!b]	
	\begin{center}	
		\includegraphics[width=0.95\linewidth, trim={0cm, 0cm, 0cm, 0cm, 0mm}, clip=true]{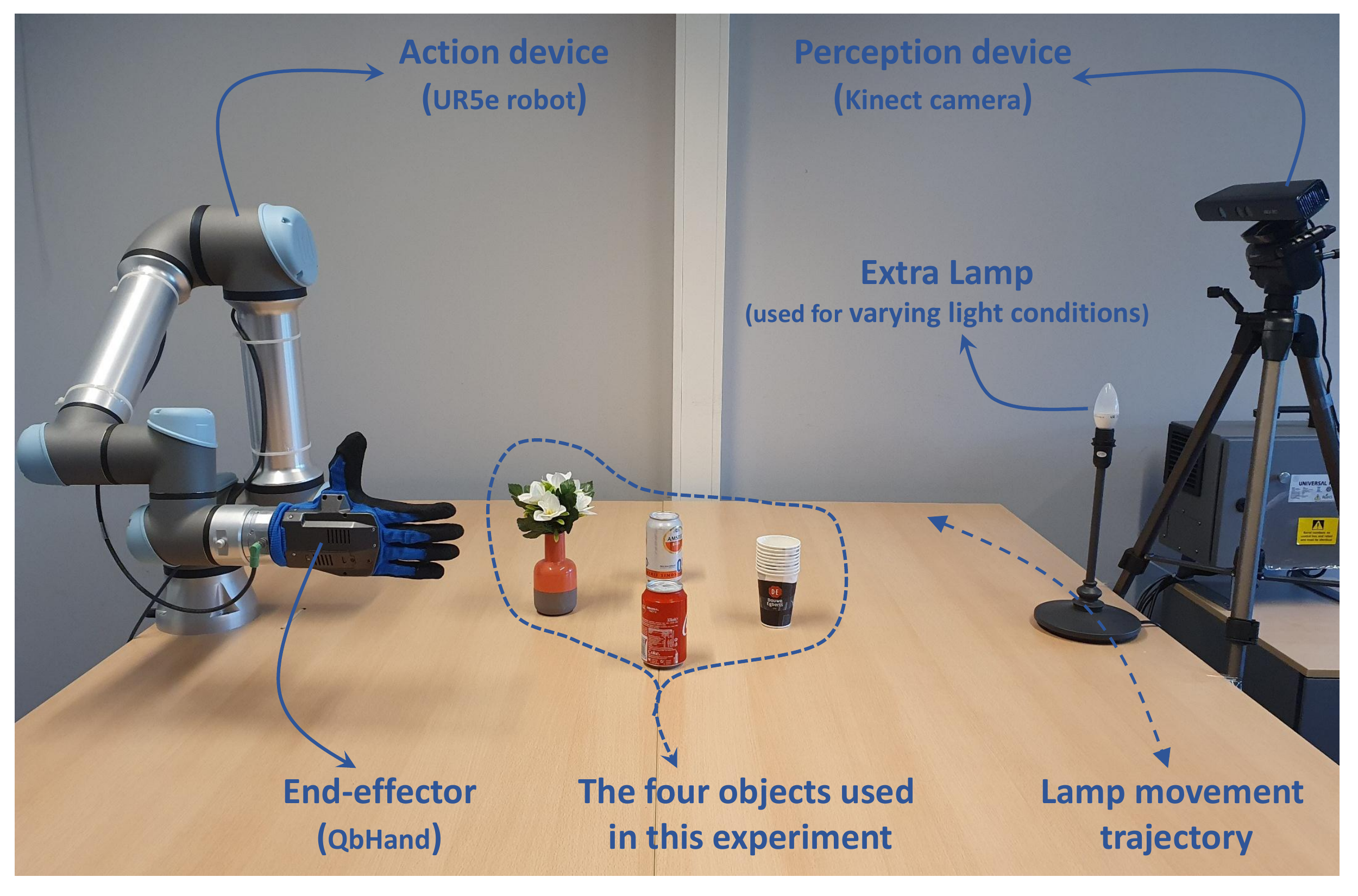}
	\end{center}
	\caption{Our experimental setup consists of a computer for human-robot interaction purposes, a Kinect sensor, and a UR5e robotic-arm as the primary sensory-motor embodiment for perceiving and acting upon its environment. We also use an extra lamp to test the performance of the system under varying light conditions. }
	\label{experiment_setup}
	\vspace{-5mm}
\end{figure}

Fig.~\ref{GCA_NLC} shows the global classification accuracy obtained by the best combination of shape and color as a function of the number of learned categories in three different color spaces (i.e., the best configuration in each color space is highlighted by the blue color in respective tables). One important observation is that accuracy decreases in all approaches, as more categories are introduced. This is expected since a higher number of categories known by the system tends to make the classification task more difficult. 

\subsection{Real-robot experiment}
To show the strength of the proposed approach, we carried out a real-robot experiment in the context of the \textit{serve\_a\_coke} scenario. We have integrated the proposed approach into the cognitive robotics system presented in \cite{kasaei2019interactive}. In this experiment, a table is in front of a Kinect sensor, and a user interacts with the system. There are one instance of four object categories on the table: \textit{CokeCan}, \textit{BeerCan}, \textit{Cup} and \textit{Vase}. This is a suitable set of objects for this test, since there are objects with very similar shapes and different colors (\textit{CokeCan}, \textit{BeerCan} and \textit{Cup}) and also objects with very different shapes and similar colors (\textit{CokeCan} and \textit{Vase}). The experimental setup is shown in Fig.~\ref{experiment_setup}. It consists of a computer for human-robot interactions, a Kinect sensor for perceiving the environment, and a Universal Robot (UR5e) for manipulation purposes.

\begin{figure}[!b]
	\centering
	\includegraphics[width=0.95\linewidth, trim={0cm, 0cm, 0cm, 0cm, 0mm}, clip=true]{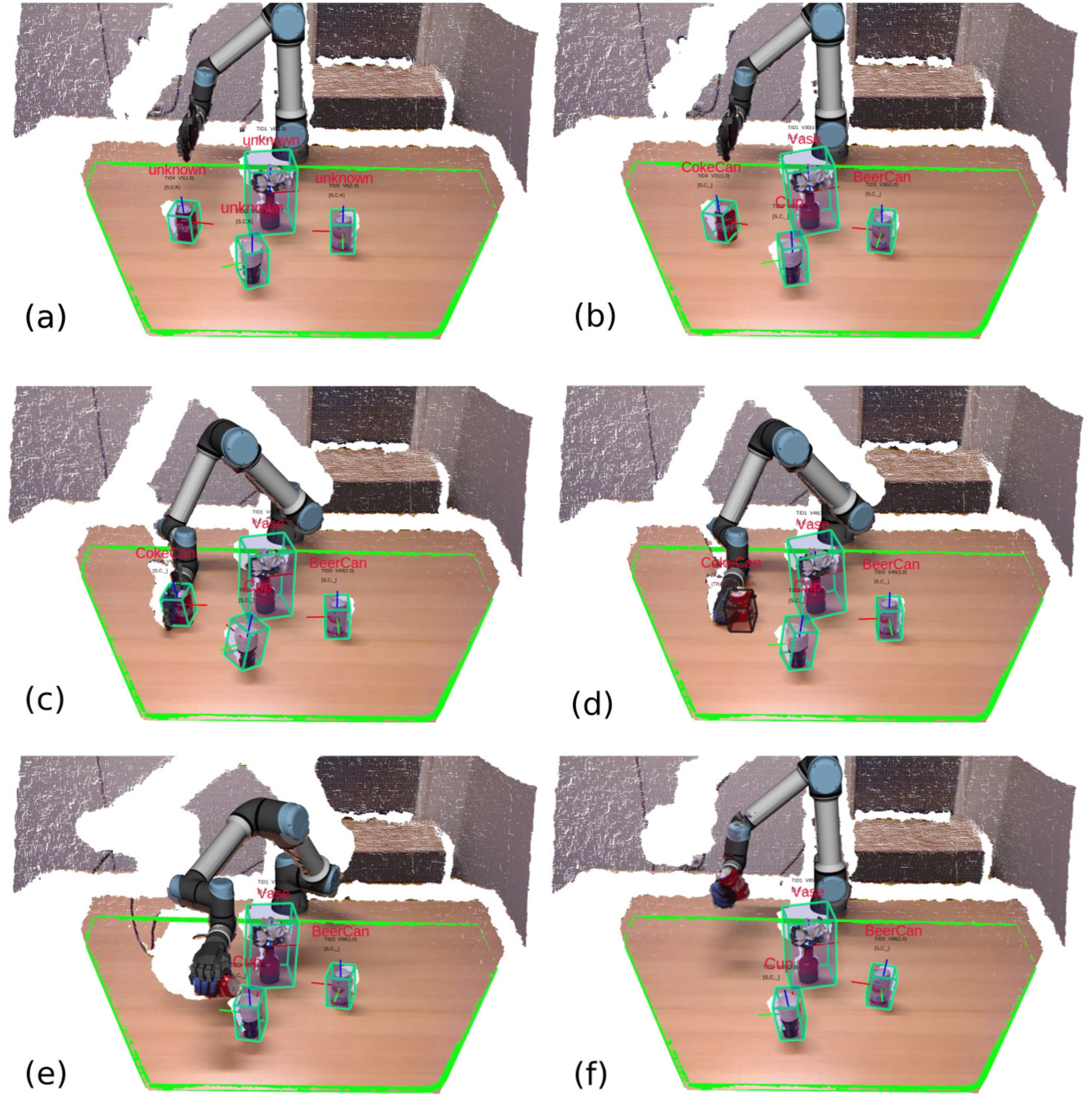}
	\caption{System performance during the \textit{serve\_a\_coke} scenario; (\textit{a}) Initially, the system starts with no knowledge of any object. The posture of the UR5e arm in each state is also visualized. The table is then detected, as shown by the green polygon. Afterward, the object candidates are detected and highlighted by different bounding boxes. The local reference frame of each object represents the pose of the object as estimated by the object tracking module. (\textit{b}) A user then teaches all the active objects to the system, and all objects are correctly recognized, i.e., the output of object recognition is shown in red on top of each object. (\textit{c}) The robot then finds out the \textit{CokeCan} object and goes to its pre-grasp area and (\textit{d}) picks it up first from the table. (\textit{e}) The robot retrieves the position of \textit{Cup} first, and then moves the \textit{CokeCan} on top of the \textit{Cup} and serves the drink. (\textit{f}) Finally, the robot goes back to the initial position.}
\label{setup_real_robot}
\end{figure}

Fig.~\ref{setup_real_robot} presents some snapshots of this experiment.  {It is worth mentioning that a constraint has been applied to the Z-axis of objects, which forces its initial direction to be similar to the direction of the table's Z-axis.} At the start of the experiment, the set of categories known to the system is empty, and therefore, the system recognizes all table-top objects as \textit{Unknown} (Fig.~\ref{setup_real_robot} (\textit{a})). {It should be noted that the robot can learn about a set of categories in advance from batch data (i.e. dataset of observations with ground truth labels), and improves its knowledge in active and on-line manners.} A user interacts with the system by teaching all object categories. The system conceptualizes them using the extracted object views and recognizes all objects properly (Fig.~\ref{setup_real_robot} (\textit{b})). In this task, the robot must be able to detect the pose of objects as well as recognize the label of all active objects. Afterward, it has to grasp the \textit{CokeCan} object (Fig.~\ref{setup_real_robot} (\textit{c, d})) and transport it on top of the \textit{Cup} object and serve the drink (Fig.~\ref{setup_real_robot} (\textit{e})). The robot finally returns to the initial pose (Fig.~\ref{setup_real_robot} (\textit{f})).
It was observed that the proposed object descriptor is capable to provide distinctive global feature for recognizing geometrically similar objects with different color and vise versa. This evaluation also illustrates the process of learning object categories in an open-ended fashion.  %A video of this session is also available online at: \href{https://youtu.be/eNdIMWj9ido}{{https://youtu.be/eNdIMWj9ido}}

\begin{figure}[!t]
	\centering
	\includegraphics[width=0.95\linewidth, trim={0cm, 0cm, 0cm, 0cm, 0mm}, clip=true]{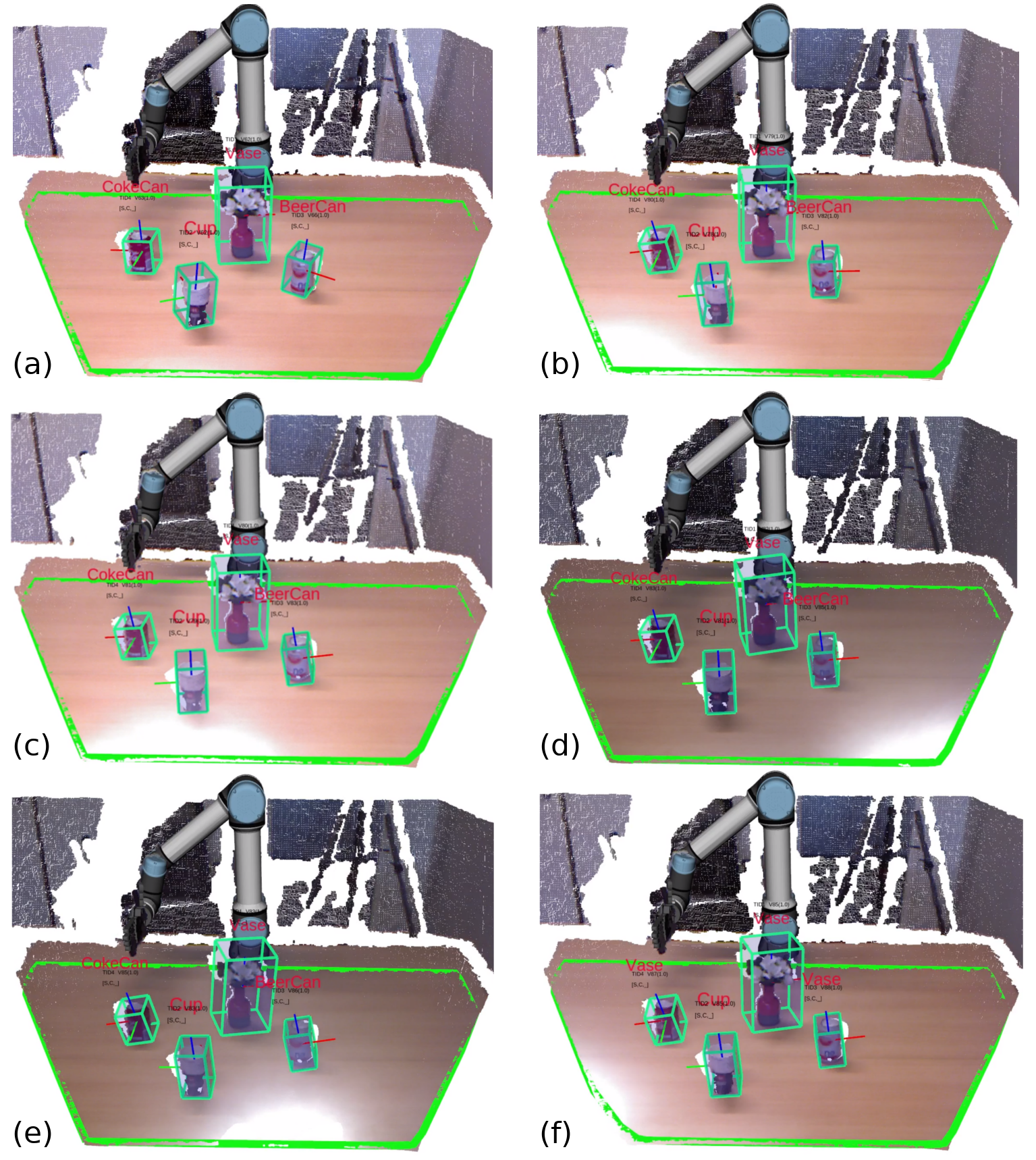}
	\caption{{System performance under various lighting conditions: (a -- e) the robot could recognize all objects correctly under varying light conditions; (f) some misclassifications also happened where the robot could not recognize \textit{BeerCan} and \textit{CokeCan} objects correctly.}}
\label{varing_light}
\end{figure}

{Most of object recognition systems are sensitive to lighting conditions and require a fair amount of time and effort to calibrate. We performed another real-time demonstration to evaluate the robustness of the proposed approach under various lighting conditions. As shown in Fig.~\ref{experiment_setup}, we placed a lamp underneath the Kinect camera as an extra source of illumination and changed its position through the experiment. Figure~\ref{varing_light} shows a sequence of snapshots of the behavior of the system in this experiment. It was observed that by combining shape features and color constancy information, we achieved a good level of robustness against varying lighting condition. In particular, the system could recognize all objects correctly under various lighting conditions (see Fig.~\ref{varing_light} \textit{a} -- \textit{e}). Some misclassification also happened. Figure~\ref{varing_light} (\textit{f}) shows a snapshot where the robot incorrectly recognized the \textit{BeerCan} and the \textit{CokeCan} objects as \textit{Vase}, and correctly recognized the \textit{Vase} and the \textit{Cup} objects.  A video of these experiments is available online at: \href{https://youtu.be/d7SRt_xGONw}{{https://youtu.be/d7SRt\_xGONw}}}

\section {Conclusion}
\label{conclusion}
In this article, we have investigated the importance of shape features, color constancy information, and similarity measures in open-ended 3D object recognition. Towards this goal, an instance-based 3D object category learning and recognition has been developed, which can be seen as a combination of a memory system, an object representation, a similarity measure, and the nearest neighbor classifier. We have selected three state-of-the-art global 3D shape descriptors, namely GOOD~\cite{kasaei_good:_2016}, ESF~\cite{ESF}, and VFH~\cite{VFH}, which provide a good trade-off between descriptiveness, computation time and memory usage and are suitable for real-time robotic application. Besides, a multitude of distance functions has been implemented to measure the similarity of two object views. Accordingly, $224$ system configurations have been examined in offline settings.

The offline experiments have been performed to optimize the parameters of selected shape descriptors and investigate the importance of similarity measures. It was observed that the combination of the GOOD descriptor ($number\_of\_bins = 15$) and $Manhattan$ function made the best result in terms of both accuracy and computation time. We then investigate the importance of color information in an open-ended learning setting. In particular, we have added the color constancy information of an object to its shape description. A set of $330$ open-ended experiments has been performed in three popular color spaces including: RGB, YUV, and HSV. In this round of experiments, we adopted a teaching protocol to incrementally evaluate the performance of the system concerning several characteristics, including descriptiveness, scalability, and experiment time. 

Experimental results show that the overall classification performance obtained with the proposed shape+color approach is clearly better than the best accuracies achieved with the \textit{color-only} and \textit{shape-only} methods. In particular, by setting the color weight parameter in the range of $0.6\le w \le0.9$ in all color spaces, the agent could learn all categories in all experiments with stable performance. This might suggest that there are reliable color differences between categories and similar color values within categories in the Washington RGB-D dataset~\cite{RGBD_dataset}. This is not always the case in the real-world environment. Furthermore, it was observed that the performance of the agent with color-only setting ($w = 1$) was better than the shape-only configuration ($w = 0$). This might be caused by a data bias in the dataset. Concerning computational time (QCI), the best result was obtained with shape+HSV ($w = 0.9$), followed by the shape+YUV with the same $w$. It was also observed that the agent could learn new categories from very few examples in an incremental and open-ended manner. A real demonstration was also carried out to show the usefulness of the proposed method.

Although the addition of color information to the object representation improved the performance of object recognition, the number of bins representing the color constancy information was greatly outnumbered by the number of bins dedicated to the shape of the objects. The color information had a small role in the resulting histogram since only the color constancy of the object was used. In the continuation of this work, we would like to investigate the possibility of integrating color information in a concrete manner. Furthermore, separate distance functions could be used to estimate the similarity of objects in terms of shape and color information. 

\bibliographystyle{spmpsci}
\bibliography{CoR}

%\begin{acknowledgements}
%If you'd like to thank anyone, place your comments here
%and remove the percent signs.
%\end{acknowledgements}

\end{document}